\pdfoutput=1

\documentclass[11pt]{article}

\usepackage{acl}

\usepackage{times}
\usepackage{latexsym}

\usepackage{booktabs} 
\usepackage{amsmath}
\usepackage{CJKutf8}
\usepackage{makecell}
\usepackage{bbm}
\usepackage{graphicx}
\usepackage{enumitem}

\usepackage[T1]{fontenc}

\usepackage[utf8]{inputenc}

\usepackage{microtype}

\title{Utterance Rewriting with Contrastive Learning in Multi-turn Dialogue}
\author{
    Zhihao Wang\textsuperscript{\rm 1, 2}
    Tangjian Duan\textsuperscript{\rm 2}
    Zihao Wang\textsuperscript{\rm 2}
    Minghui Yang\textsuperscript{\rm 2}
    Zujie Wen\textsuperscript{\rm 2}
    Yongliang Wang\textsuperscript{\rm 2} \\
    \textsuperscript{\rm 1} Carnegie Mellon University\\
    \texttt{zhihaow2@andrew.cmu.edu} \\
    \textsuperscript{\rm 2} ANT GROUP \\
    \texttt{\{tangjian.dtj, xiaohao.wzh, minghui.ymh, zujie.wzj, yongliang.wyl\}} \\
    \texttt{@antgroup.com} \\

}

\begin{document}
\maketitle
\begin{abstract}
Context modeling plays a significant role in building multi-turn dialogue systems. In order to make full use of context information, systems can use Incomplete Utterance Rewriting(IUR) methods to simplify the multi-turn dialogue into single-turn by merging current utterance and context information into a self-contained utterance. However, previous approaches ignore the intent consistency between the original query and rewritten query. The detection of omitted or coreferred locations in the original query can be further improved. In this paper, we introduce contrastive learning and multi-task learning to jointly model the problem. Our method benefits from carefully designed self-supervised objectives, which act as auxiliary tasks to capture semantics at both sentence-level and token-level. The experiments show that our proposed model achieves state-of-the-art performance on several public datasets.
\end{abstract}

\section{Introduction}
With the development of single-turn dialogue modeling, remarkable progress has been achieved \cite{zhang2021advances} in both question answering and open-domain response generation. However, in daily dialogues, users tend to omit or refer back to avoid repetitions. Research \cite{su-etal-2019-improving} has shown that coreference and omission phenomenon exist in 33.5\% and 52.7\% of utterances, respectively, in pro-drop languages such as Chinese. Because users are capable of completing the simplified utterance by remembering the conversational history. Similarly, to equip the dialogue system with conversational memory, we would track the important history information through Dialogue States Tracking (DST). However, DST has three main problems: a) The volume of stored information is limited in long conversations. b) The stored information is pruned to avoid the redundancy of information. c) Other modules can rarely utilize the stored information in dialogue systems. Thus, To avoid problems of DST and boost the performance of dialogue understanding, recent researches propose simplifying multi-turn dialogue modeling into a single-turn problem by \textbf{I}ncomplete \textbf{U}tterance \textbf{R}erewring (\textbf{IUR}) \cite{kumar-joshi-2016-non, pan-etal-2019-improving, su-etal-2019-improving, elgohary-etal-2019-unpack, zhou-etal-2019-unsupervised, liu-etal-2020-incomplete, zhou-etal-2019-unsupervised, liu-etal-2020-incomplete}. Specifically, IUR is expected to recovering coreferred and omitted mentions in an incomplete utterance.

The examples from Table \ref{tb:examples} correspond to the phenomena of coreference and omission respectively. "{\begin{CJK}{UTF8}{gbsn}他\end{CJK}}"(he) from User\_query is a coreference to "{\begin{CJK}{UTF8}{gbsn}周杰伦\end{CJK}}"(Jay-Chou) in the first example and User\_query from the second example omits the subject "{\begin{CJK}{UTF8}{gbsn}上海\end{CJK}}"(Shanghai). In the two examples from Table \ref{tb:examples}, each User\_query will be rewritten to User\_query* by IUR. The dialogue system could modeling rewritten utterances more precisely without considering previous utterances. Besides, IUR is an extensible module that could be effortlessly integrated into different stages of a dialogue system, such as intent recognition or question answering tasks.

\begin{small}
\begin{table*}[htb]
\large
\centering
\begin{tabular}{c|c}
\toprule
\textbf{Turn}  & \textbf{Utterance}\textit{(Translation)} \\
\midrule
User\_context: &  \makecell[c]{{\begin{CJK}{UTF8}{gbsn} 你喜欢周杰伦吗 \end{CJK}} \\ \textit{Do you like Jay-Chou}} \\

System\_context: & \makecell[c]{{\begin{CJK}{UTF8}{gbsn} 我喜欢周杰伦 \end{CJK}} \\ \textit{I like Jay-Chou}} \\

User\_query: &  \makecell[c]{{\begin{CJK}{UTF8}{gbsn} 你喜欢他哪首歌 \end{CJK}} \\ \textit{Which song of him do you like}} \\

User\_query*: &  \makecell[c]{{\begin{CJK}{UTF8}{gbsn} 你喜欢周杰伦哪首歌 \end{CJK}} \\ \textit{Which song of Jay Chou do you like}} \\
\midrule
User\_context: &  \makecell[c]{{\begin{CJK}{UTF8}{gbsn} 上海今天下雨吗 \end{CJK}} \\ \textit{Does it rain in Shanghai today}} \\

System\_context: & \makecell[c]{{\begin{CJK}{UTF8}{gbsn} 上海今天下雨 \end{CJK}} \\ \textit{It rains today in Shanghai}} \\

User\_query: &  \makecell[c]{{\begin{CJK}{UTF8}{gbsn} 为什么最近总是下雨 \end{CJK}} \\ \textit{Why does it always rain recently}} \\

User\_query*: &  \makecell[c]{{\begin{CJK}{UTF8}{gbsn} 为什么上海最近总是下雨 \end{CJK}} \\ \textit{Why does it always rain recently in Shanghai}} \\
\bottomrule
\end{tabular}
\caption{Some examples of coreference and omission in daily life.}
\label{tb:examples}
\end{table*}
\end{small}

Previous work normally designs models in a two-stage way\cite{yin-etal-2018-zero} including detecting the omitted or coreferred words and conducting the resolution task. However, it will introduce an accumulated error, i.e., a false detection leads to a false resolution. Recently, more and more researchers are focusing on designing end-to-end model architectures to solve IUR \cite{pan-etal-2019-improving}, \cite{su-etal-2019-improving}, as end-to-end models could avoid the problems of error accumulation and achieve better performance and speed. However, previous end-to-end models fail to explore the key traits of IUR task fully. For example, the natural assumption of IUR that the completed utterance should be semantically equivalent to the original dialogue is often neglected, while this trait could naturally be modeled via a contrastive learning paradigm to boost the performance of IUR. Furthermore, we can still take advantage of two-stage methods while not being harmed by error accumulation with contrastive learning. In this work, we propose three simple yet effective designs to improve the performance of current IUR models. Our contributions are as follows:

\begin{enumerate}
\item As far as we know, we are the first to introduce contrastive learning into IUR.
\item We explore the key traits of IUR and modeling them in a multi-task learning paradigm.
\item Maintaining a fast inference speed, our approach achieves state-of-the-art performance on several Chinese datasets across different domains.
\end{enumerate}

\section{Related Work}
Previous researches treat IUR as coreference resolution problem, adopting two-stage models to construct a detection-resolution pipeline \cite{yin2017deep, yin-etal-2018-zero}. However, these methods often assume available golden syntactic parse trees, which are rare in real datasets. Recently, people have paid more attention to end-to-end models. The main-stream architectures could be classified into three categories: autoregressive(generation), semi-autoregressive and and non-autoregressive(sequence tagging). Most previous work model the problem as a standard autoregressive text generation task \cite{su-etal-2019-improving, quan2019gecor, elgohary-etal-2019-unpack}. They adopt sequence-to-sequence models with copy mechanism to tackle the problem.

Besides, there are also work tackling IUR in a sequence-tagging paradigm RUN \cite{liu-etal-2020-incomplete}, RAST \cite{hao2020robust}. RUN designs the model similarly to semantic segmentation in the computer vision domain, while RAST predicts target rewritten span for each token in the original query. These works are achieving the best performance on public datasets, as they further reduce the search space. In the meantime, avoiding the generation process means that these models have fast inference speed as they do not need beam search. Naturally, semi-auto-regressive model architectures are also explored by SARG\cite{huang2020sarg}.

Previous works also try to improve performance by designing specific tasks to utilize the traits of IUR. Through the review of previous work, we conclude the specific tasks into five perspectives: Pretraining, Keywords Detection, Search Space Reduction, Intent Consistency Constraint and Sentence Fluidity Supervision. 

Since the utterance structure changes little in IUR, one can get abundant weak label training data by deleting the informative common span between a query and its context in a large raw dialogue corpus. Pretrained from these weak data is proved efficient by Teresa\cite{9413557}, Few-shot generative QA query rewriting\cite{10.1145/3397271.3401323} and many other work. PAC\cite{pan-etal-2019-improving} seeks to get additional gain in performance by imitating previous two-stage methods. The idea is to detect keywords first and then append those words to the context. However, the error accumulation is not avoided in PAC. SRL\cite{xu2020semantic} train a model of semantic role labeling to highlight the core meaning of keywords in dialogue as a kind of prior knowledge for the model. Teresa uses a rank algorithm to calculate the importance of each token and pass it to following steps. \cite{su-etal-2019-improving, hao2020robust, liu-etal-2020-incomplete} design their models in different architectures while sharing the same advantage of reduced search space and achieve good performance. One important feature of IUR is that the rewritten query must be complete and semantically equal to previous context and incomplete query. Thus, a natural idea is to add additional task to push the model to follow this trait. CREAD\cite{tseng2021cread} adopts a binary classification task to decide whether the original query intent is complete or not. Teresa performs an KL-Divergence loss between the original and rewritten query to force their intent to be same. Lastly, although sequence-tagging model architectures are achieving best performance these days, they face the problem of readability as they do not have a large vocabulary. RUN adds additional connection words into the context. RAST uses reinforcement learning to supervise the sentence fluidity of predictions.

As mentioned before, our method fully explores two critical traits of IUR: keywords detection and intent consistency constraint. We utilize contrastive learning and multi-task learning to avoid accumulated errors in words detection. Besides, we argue that intent consistency constraint is not fully utilized yet and models can gain more from intent consistency constraint by modeling this trait in a contrastive learning way.

\section{Methodology}
\subsection{Task Definition}
Here we give the formal definition of IUR. Given the dialogue history and current utterance as $(H,U_n)$, $H=(U_1,U_2,...,U_n)$ is the history utterances of the dialogue. The target of IUR is to learn a function to rewrite $U_n$ to $R$: $f(H,U_n) \to R$. We need to notice that $R$ and $(H,U_n)$ are semantically equivalent and $R$ is self-contained, i.e. $R$ could be understood without context.
 
\subsection{Baseline}
\begin{figure*}[htb]
\large
\centering
\includegraphics[width=0.6\textwidth, height=0.28\textwidth]{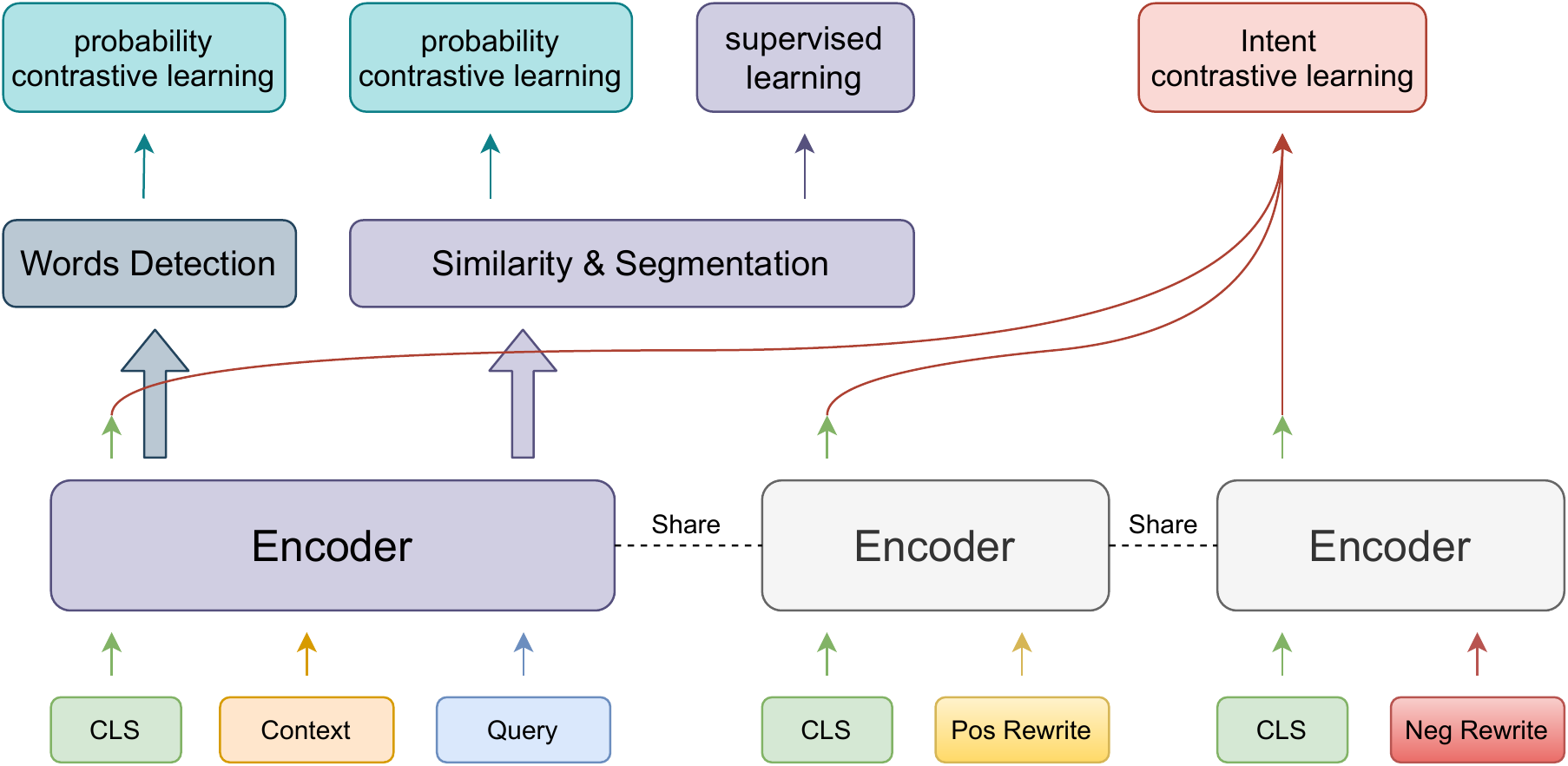} 
\caption{The framework of our proposed model. Purple cells are main modules of baseline model (RUN) and other cells are our constructed multi-tasks.}
\label{pic:MAINMODEL}
\end{figure*}

Our baseline is based on the current state-of-the-art model RUN \cite{liu-etal-2020-incomplete}, so we first give a brief introduction to RUN. The main modules of RUN are shown in purple cells in Figure \ref{pic:MAINMODEL}. The model is defined as:
\begin{equation}
\label{eq:pmat}
P_{mat} = f(CQ)
\end{equation}
Specifically, $CQ$ is the input that denotes concatenated history context $C$ and current utterance $Q$. The model learns a mapping function $f$ to predict from $CQ$ to the word-level edit operation matrix $P_{mat} \in R^{M \times N}$.

The objective function is defined as: 
\begin{equation}
\label{eq:lossmat}
L_{mat} = \frac{1}{M\times N}\sum_{i=0}^{M \times N}CE(P_{mat}^{i},Y_{mat}^{i})
\end{equation}

where $Y_{mat}^{i}$ is the target edit operation of pixel-level sample $i$. $CE$ is the notation of cross-entropy loss. 


\subsection{Words Detection}
Our first additional task is keywords detection. This task is inspired by two-stage methods of equipping the encoder with the capability of detecting coreferred words while not introducing accumulated error.

\begin{figure}[htb]
\large
\centering
\includegraphics[scale=0.38]{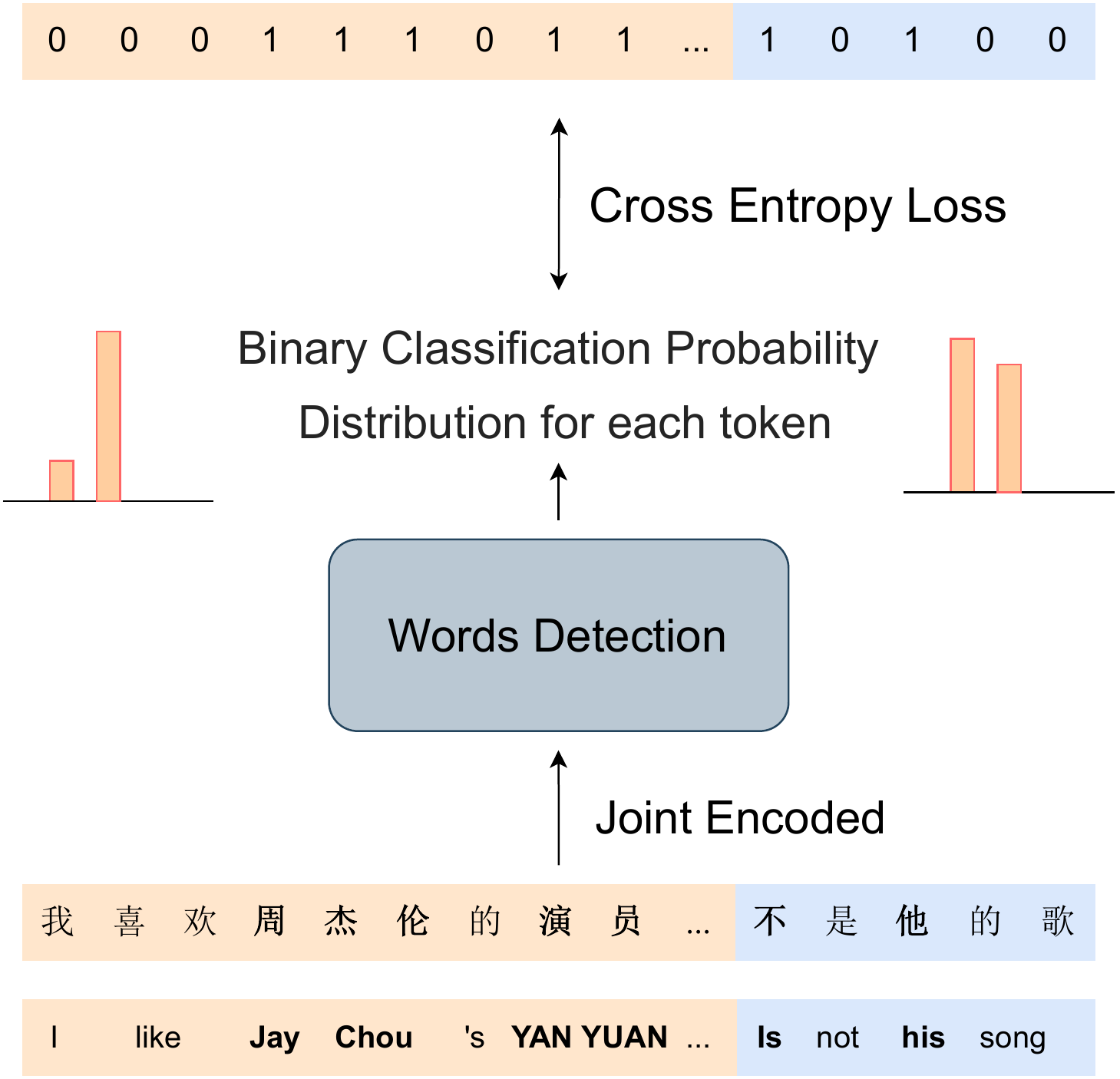}
\caption{Word Detector}
\label{pic:WORDDETECTOR}
\end{figure}

As shown in Figure \ref{pic:WORDDETECTOR}, we add a \textbf{W}ords \textbf{D}etection module (WD) on top of the encoder, which performs a binary classification task for each token. WD is composed of a one-layer feed-forward neural network and takes the hidden states of joint encoded $CQ \in R^{(M + N) \times H}$ as input. The output is a binary probability distribution $P_{detect} \in R^{(M+N) \times 2}$. This is a sequence tagging problem in essence and the loss function of the word detector is as follows: 
\begin{equation}\label{eq:lossdet}
L_{det} = \frac{1}{M+N}\sum_{i=0}^{M+N}CE(P_{det}^{i},Y_{det}^{i})
\end{equation}
$Y_{det}^{i}$ is the golden keywords label of sample $i$. The target of word detector is to minimize the average cross-entropy between predictions and labels. Due to the nature of modeling IUR as a sequence tagging problem (predicting edit operations for each token), we can easily assign the detection label for each token which requires no additional labeling resources. An example is shown in Figure \ref{pic:WORDDETECTOR}, "{\begin{CJK}{UTF8}{gbsn}演员\end{CJK}}"(YAN YUAN) and "{\begin{CJK}{UTF8}{gbsn}周杰伦\end{CJK}}"(Jay Chou) from context are keywords as they are the omitted and the coreferred nouns. "{\begin{CJK}{UTF8}{gbsn}不\end{CJK}}"(Is) and "{\begin{CJK}{UTF8}{gbsn}他\end{CJK}}"(his) from current utterance are keywords as they represent the positions where omission and coreference occur.

\subsection{Intent Consistency Constraint via Contrastive Learning}
\textbf{I}ntent \textbf{C}onsistency \textbf{C}onstraint (ICC)\cite{9413557} is important in IUR as the rewritten query is supposed to be consistent with the contextual query in the intent space. In the meantime, contrastive learning aims to learn effective representation by pulling semantically close neighbors together and pushing apart non-neighbors \cite{1640964}. We observe that the definition of ICC is naturally aligned with the purpose of contrastive learning. Thus, we introduce \textbf{C}ontrastive \textbf{L}earning based \textbf{I}ntent \textbf{C}onsistency \textbf{C}onstraint(CLICC) to the model. The objective of CLICC is to pull close the intent of joint $CQ$ and gold rewrite $R$ but push apart the intent of joint $CQ$ and incomplete queries. The steps are as follows:

\begin{enumerate}[wide, labelwidth=!, labelindent=0pt]
\item[\textbf{Anchor}] As shown in Figure \ref{pic:MAINMODEL}, we insert a ``[CLS]" token before the concatenated context and query. The hidden states of ``[CLS]" token represents the intent of joint $CQ$.

\item[\textbf{Positive Instance}] According to the prior assumption of IUR, a gold rewrite $R$ is supposed to be consistent with the contextual query in intent dimension, so we use $R$ as positive instance. We first insert a ``[CLS]" token before the gold rewrite $R$ and then feed $R$ into the same encoder to get the encoded hidden states of ``[CLS]" in $R$.

\begin{figure}[htb]
\large
\centering
\includegraphics[width=0.45\textwidth, height=0.25\textwidth]{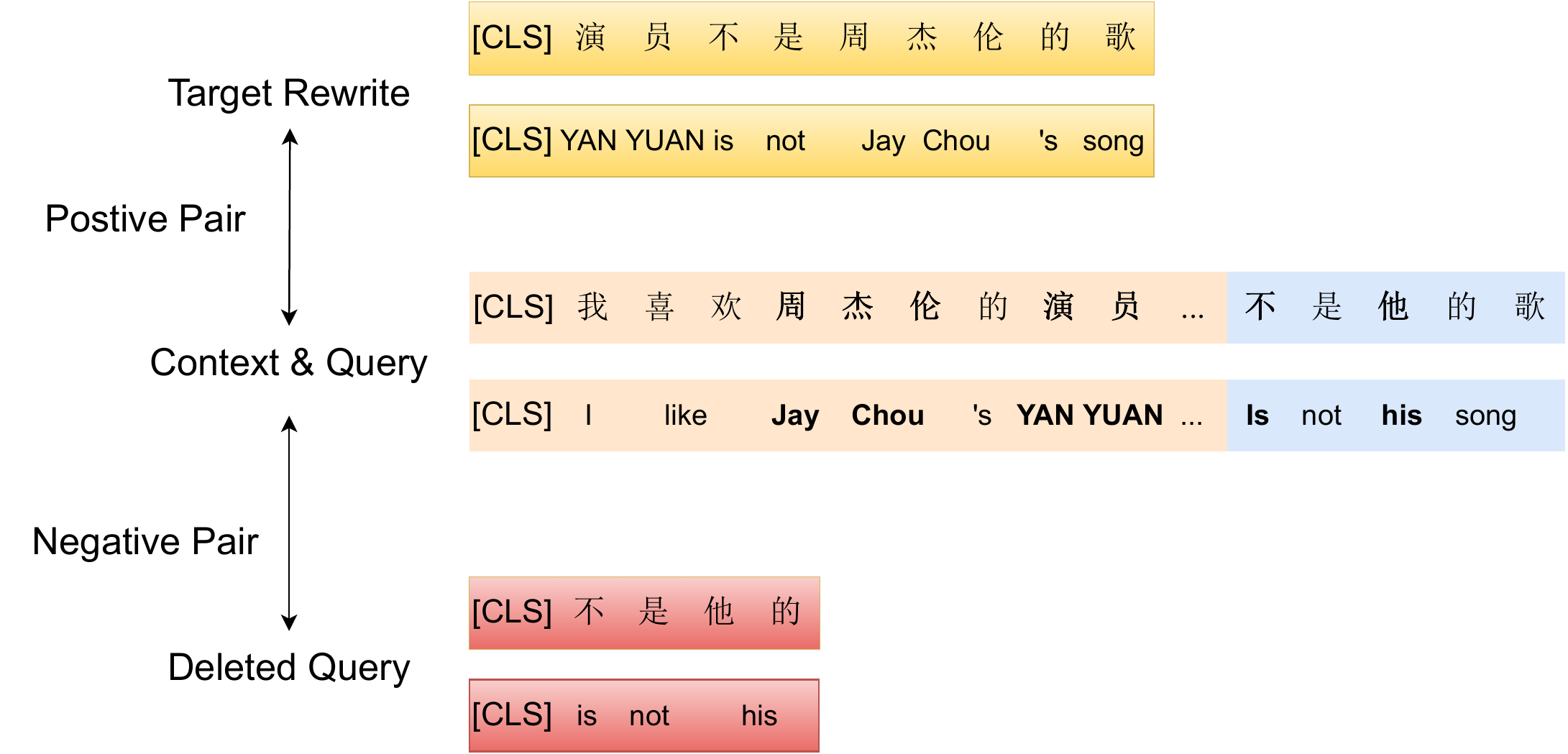}
\caption{An example of query needs rewriting. When the intent of original query is not complete, the deletion could further corrupt the query into a more incomplete one.}
\label{pic:INTENT}
\end{figure}

\begin{figure}[!h]
\large
\centering
\includegraphics[width=0.45\textwidth, height=0.25\textwidth]{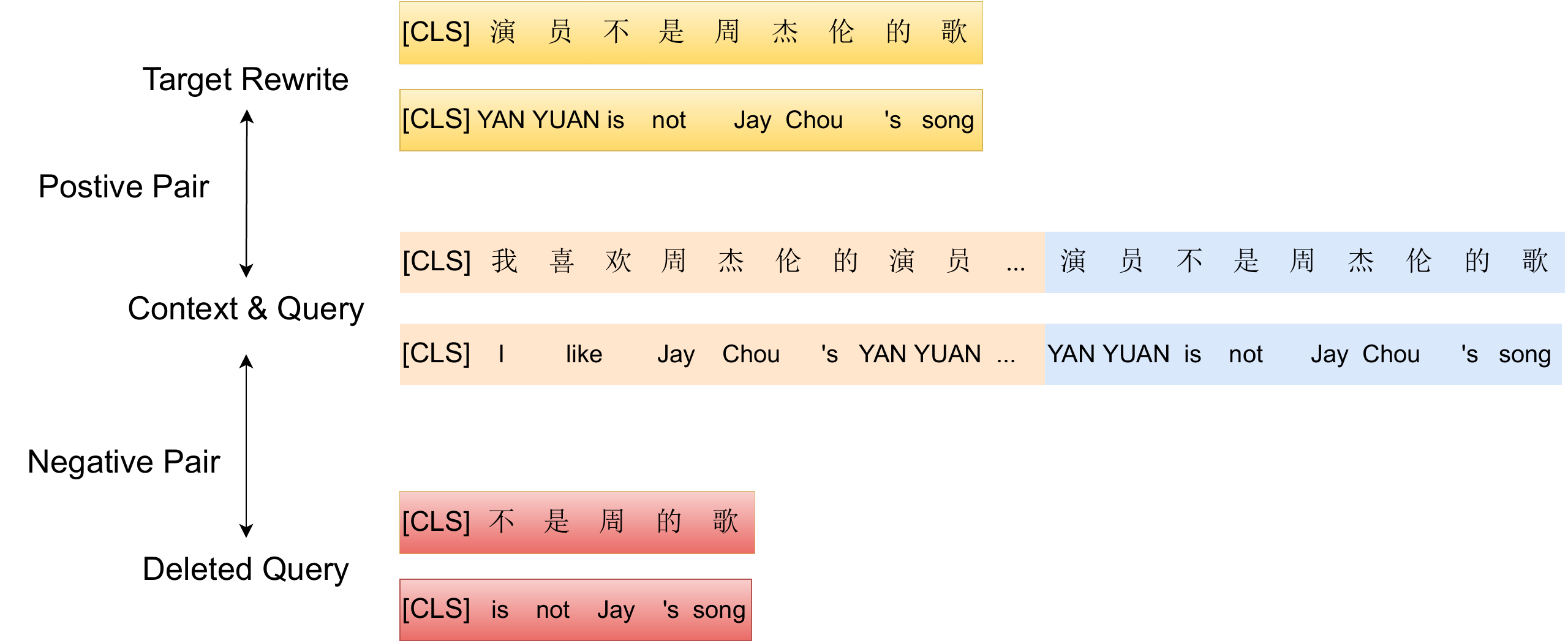}
\caption{An example of query does not need rewriting. When the intent of original query is complete, the deletion could corrupt the query into an incomplete one.}
\label{pic:INTENTNONEED}
\end{figure}

\item[\textbf{Hard Negative Instances}] For the construction of hard negative instances, we adopt a simple yet effective strategy: Random Token Deletion. We randomly delete 20\% tokens in the original query for each sample. Then we insert a ``[CLS]" token before the deleted query and feed it through the same encoder to get the intent representation of the deleted query. Because of similar sentence structure, these instances are hard to be distinguished from the original query. This strategy is effective in three ways:

\begin{enumerate}
\item Token deletion ensures a complex negative instance for each sample even if the original queries are complete. 

\item Token deletion for incomplete original queries lowers the learning difficulty for the model in beginning epochs, as this strategy creates a more severely incomplete query, which is farther from the complete query.

\item Random token deletion enriches the negative instances as the deleted queries for a certain sample $i$ are different between epochs. 
\end{enumerate}

As shown in Figure \ref{pic:INTENT}, the original query is incomplete and random token deletion further corrupts the query. As shown in Figure \ref{pic:INTENTNONEED}, random token deletion converts the complete query into an incomplete one.

\item[\textbf{Easy Negative Instances}] We also fill the negative sample space with abundant, accessible negative instances. For a certain sample $i$ in one batch, other positive and hard negative instances will be treated as negative pairs for sample $i$, as their intent is naturally different. 

\item[\textbf{Loss function}] Following \cite{chen2020simple}, we adopt the normalized temperature-scaled cross-entropy loss (NT-Xent) as the contrastive objective for CLICC. Suppose we have $N$ randomly sampled dialogues from the training set as a mini-batch during each training step. There are $N$ representations for all positive instances, negative instances and anchors in the batch. The objective function is slightly modified to train each anchor to find its counterpart among $3N-2$ in-batch negative samples: 
\begin{gather}
\label{eq:lossicon}
L_{icon} =-log \frac{exp(sim(r_i,r_i^{+})/\tau)}{Anchor + Pos + Neg} \\ 
Anchor =\sum_{j=1}^{N}\mathbbm{I}_{[j \ne i]}exp(sim(r_i,r_j)/\tau) \\
Pos =\sum_{j=1}^{N}\mathbbm{I}_{[j \ne i]}exp(sim(r_i,r_j^{+})/\tau) \\
Neg =\sum_{j=1}^{N}exp(sim(r_i,r_j^{-})/\tau)  
\end{gather}

where sim(·) is defined as the cosine similarity function, $\tau$ in temperature parameter and $\mathbbm{I}$ is the indicator function. $icon$ is the short notation of \textbf{I}ntent \textbf{Con}trastive.

\end{enumerate}

\subsection{Probability Space Contrastive Learning}
\begin{figure}[htb]
\large
\centering
\includegraphics[scale=0.3]{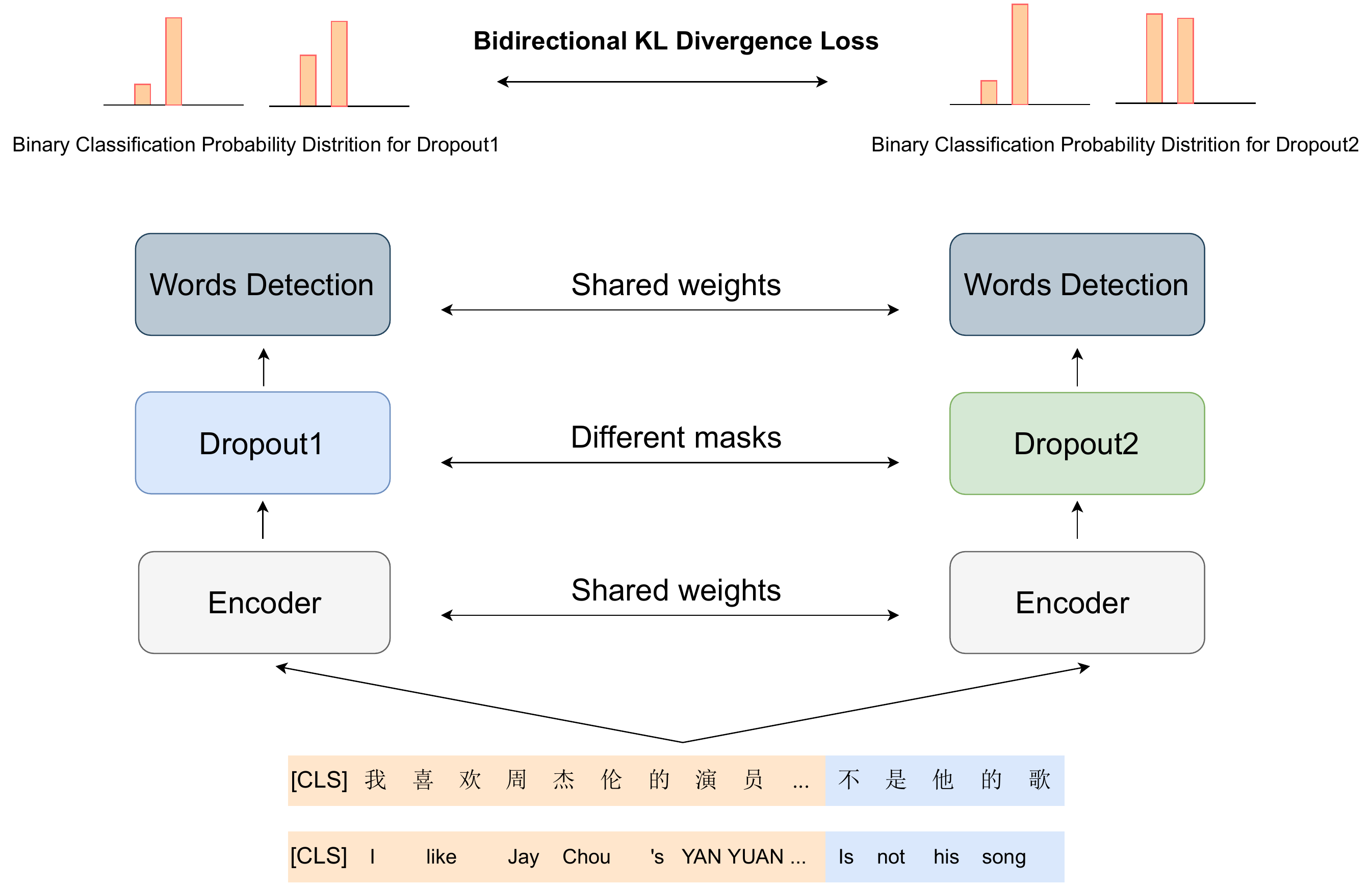}
\caption{PCL on word detection probability distribution.}
\label{pic:PROBWD}
\end{figure}

\begin{figure}[htb]
\large
\centering
\includegraphics[scale=0.3]{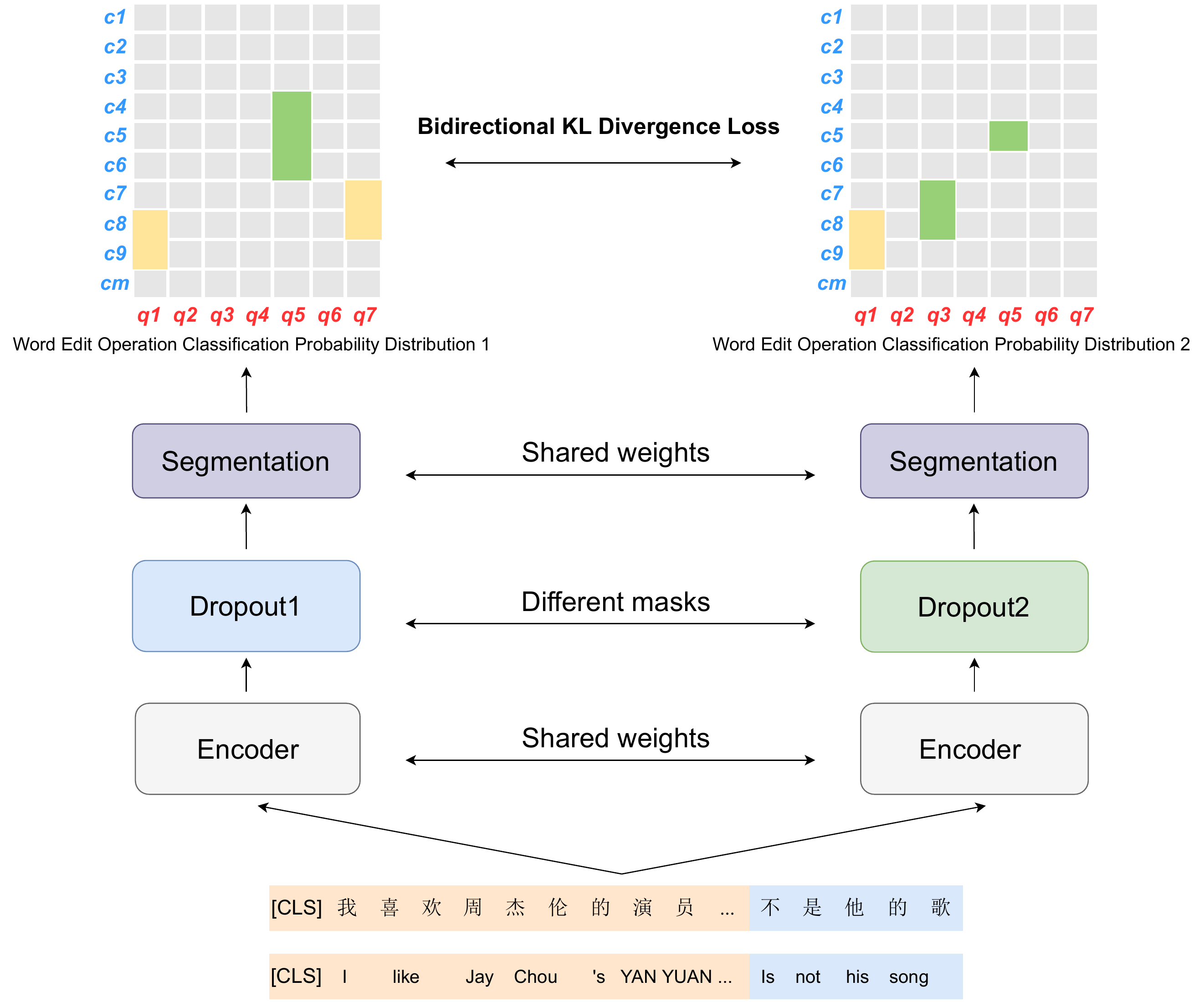}
\caption{PCL on word edit operation probability distribution}
\label{pic:PROBMATRIX}
\end{figure}

Inspired by the effective result of CLICC, we introduce \textbf{P}robability \textbf{C}ontrastive \textbf{L}earning (\textbf{PCL}) modules on top of both words detection and semantic segmentation modules to help the optimization in learning. This target target is to make the predicted probability distribution closer for a positive pair. The steps are as follows:

\begin{enumerate}[wide, labelwidth=!, labelindent=0pt]

\item[\textbf{Positive Instances}] One way to effectively create positive instances in NLP tasks is through data augmentation such as word reordering, deletion, repeating and substitution\cite{feng2021survey}. However, these augmentation are not suitable for the PCL module in IUR in two aspects: a) we may unconsciously delete or repeat the keywords in context and queries. b) We could potentially change the original intent of dialogues. These two risks both lead to a false predicted distribution. Thus, instead of above data augmentation techniques, similar to ideas in SimCSE \cite{gao2021simcse}, we use Dropout\cite{JMLR:v15:srivastava14a} to safely acquire the positive instances. Specifically, we feed the same dialogue utterances to the encoder and embedder twice to get two different dropped context and query $CQ_1$,$CQ_2$. We use $CQ_1$ as the anchor and $CQ_2$ and the positive instance.

\item[\textbf{Loss function}] As shown in Figure \ref{pic:PROBWD} and Figure \ref{pic:PROBMATRIX}, we constrain the predicted distributions of WD and word-level edit matrix of a positive pair to be close by minimizing the bidirectional Kullback-Leibler(Bi-KL) divergence loss between predicted distributions as follows: 
\begin{equation}\label{eq:losspcon}
    L_{pcon}= \frac{1}{2}(L_{det} + L_{mat})
\end{equation}
where $pcon$ is notation of \textbf{p}robability, $L_{det}$ and $L_{mat}$ denote Bi-KL divergence loss of words detection and edit matrix respectively:
\begin{gather}\label{equ4}
L_{det}=KL(P_{det}^{CQ_1}||P_{det}^{CQ_2})+KL(P_{det}^{CQ_2}||P_{det}^{CQ_1}) \\
L_{mat}=KL(P_{mat}^{CQ_1}||P_{mat}^{CQ_2})+KL(P_{mat}^{CQ_2}||P_{mat}^{CQ_1}) 
\end{gather}
where $det$ and $mat$ are the notation of \textbf{det}ection and \textbf{mat}rix respectively. $KL$ is Bi-KL divergence loss.

\item[\textbf{Negative Instances}] We do not need negative instances in PCL as we are not being exposed to the risk of model collapse compared to representation contrastive learning. Model collapse is the extreme opposite of Uniformity explained in \cite{wang2020understanding}, which means all representations of data are centering to one point in the hypersphere. This is avoided in PCL because though bidirectional KL is pulling two distributions closer because the original supervised target ensures the correct optimization direction.
\end{enumerate}

\subsection{Final Learning objectives}
Finally, we combine all tasks together and train them simultaneously by taking the weighted summation of all loss functions and the final loss function is shown as Equation \ref{eq:lossfinal}
\begin{equation}
\label{eq:lossfinal}
\begin{split}
L_{forward} &= L_{mat}^{CQ_1} +L_{mat}^{CQ_2}+\alpha(L_{det}^{CQ_1}+L_{det}^{CQ_2})\\
L_{final} &= L_{forward} + \beta(L_{icon}) + \gamma(L_{pcon}) \\
\end{split}
\end{equation}
where $\alpha,\beta,\gamma$ are coefficients for three introduced addition learning tasks, WD, CLICC and PCL.

\section{Experiments}
In this section, we conduct thorough experiments to demonstrate the effectiveness of our approach.
 
\subsection{Datasets}
We conduct experiments on two Chinese public datasets in open-domain dialogues: MULTI \cite{pan-etal-2019-improving} and REWRITE \cite{su-etal-2019-improving}. We use the same data split method for these datasets as their original paper. We display the statistics of two datasets in Table \ref{tb:datasetstatistic}. 

\begin{table}[htb]
\large
\centering
\begin{tabular}{c|cc}
\toprule
  & MULTI & REWRITE \\
\midrule
Train & 194$K$ & 18$K$ \\
Dev & 5$K$ & 2$K$ \\
Test & 5$K$ & N/A \\
Avg. C len & 25.5 & 17.7 \\
Avg. Q len & 8.6 & 6.5 \\
Avg. R len & 12.4 & 10.5 \\
\bottomrule
\end{tabular}
\caption{Statistics of the datasets. NA means the we use development set as test set. "Avg" is short for average, "C" for context, "Q" for current query, "R" for rewritten query.}
\label{tb:datasetstatistic}
\end{table}

\subsection{Baselines}
To prove the effectiveness of our approach, we take the State-of-the-art models as strong baselines including SRL\cite{xu2020semantic}, SARG\cite{huang2020sarg}, PAC\cite{pan-etal-2019-improving}, RAST\cite{hao2020robust}, T-Ptr-$\lambda$ \cite{su-etal-2019-improving} and RUN\cite{liu-etal-2020-incomplete}. 

\subsection{Evaluation Metrics}
We take automatic metrics to evaluate our approach. Following \cite{pan-etal-2019-improving}, we employ the widely used automatic metrics BLEU\cite{papineni-etal-2002-bleu}, ROUGE\cite{lin-2004-rouge}, Exact-Match(EM) and Rewriting F-score\cite{pan-etal-2019-improving}. (i) \textbf{BLEU}$_n$ (\textbf{B}$_n$) evaluates how similar the rewritten utterances are to the golden ones via the cumulative n-gram BLEU score. (ii) \textbf{ROUGE}$_n$ (\textbf{R}$_n$) measures the n-gram overlapping between the rewritten utterances and the golden ones, while \textbf{ROUGE}$_L$ (\textbf{R}$_L$) measures the longest matching sequence between them. (iii) \textbf{EM} stands for the exact match accuracy, which is the strictest evaluation metric. (iv) Rewriting \textbf{Precision}$_n$, \textbf{Recall}$_n$ and \textbf{F-score}$_n$ ($P_n,R_n,F_n$) emphasize more on how well we recover the correferred words.

\subsection{Implementation Details}
Our approach is developed based on the model architecture of RUN. We follow the original settings in RUN: The weighted cross-entropy loss is used; We used Adam\cite{kingma2017adam} to optimize the model and set the learning rate 1e-3, except for BERT\cite{devlin2019bert} as 1e-5; The embedding size and hidden size are 200 respectively. Specifically, BERT aforementioned is $BERT_{base}$. 

\subsection{Results}

\begin{table*}[htb]
\centering
\begin{tabular}{c|ccccccc}
\toprule
Model  & $F_1$ & $F_2$ & $F_3$ & \textbf{B}$_1$ & \textbf{B}$_2$ & \textbf{R}$_1$ & \textbf{R}$_2$ \\
\midrule
SRL & NA & NA & NA & 85.8 & 82.9 & 89.6 & 83.1\\
T-Ptr-$\lambda$ (n\_beam=5) & 51.0 & 40.4 & 33.3 & 90.3 & 87.7 & 90.1 & 83.0 \\ 
PAC(n\_beam=5) & 63.7 & 49.7 & 40.4 & 89.9 & 86.3 & 91.6 & 82.8 \\
SARG(n\_beam=5) & 62.3 & 52.5 & 46.4 & 91.4 & 88.9 & 91.9& 85.7 \\
RAST & NA & NA & NA & 89.7 & 88.9 & 90.9 & 84.0 \\
RUN & 69.0 & 57.1 & 48.8 & 90.7 & 87.7 & 92.0 & 85.1 \\
+WD & 70.8 & 58.2 & 49.6 & 91.1 & 88.1 & 92.1 & 85.2 \\
+CLICC & 70.2 & 57.8 & 49.3 & 91.5 & 88.6 & 92.3 & 85.7 \\
+PCL & \textbf{71.1} & \textbf{59.1} & \textbf{51.1} & \textbf{92.1} & \textbf{89.4}  & \textbf{92.6} & \textbf{86.2}\\
\bottomrule
\end{tabular}

\caption{Reuslts on Restoration-200k. All models except T-Ptr-$\lambda$ are initalized from pretrained Bert-base-Chinese model. All results are extracted from the original papers except RUN. For RUN, we reproduce the results from released code to ensure a fair comparison as we are adding modules onr RUN. The final line is the result of our complete model equipped with all three modules.}
\label{tb:multiresult}
\end{table*}

\begin{table*}[htb]
\centering
\begin{tabular}{c|cccccccccc}
\toprule
Model  & $F1$ & $F2$ & $F3$ & \textbf{EM} & \textbf{B}$_1$  & \textbf{B}$_2$  & \textbf{B}$_4$  & \textbf{R}$_1$ & \textbf{R}$_2$ & \textbf{R}$_L$ \\
\midrule
SRL & NA & NA & NA & 60.5 & 89.7 & 86.8 & 77.8 & 91.8 & 85.9 & 90.5\\
RAST & NA & NA & NA & 63.0 & 89.2 & 88.8 & 86.9 & 93.5 & 88.2 & 90.7 \\
RUN & 89.3 & 81.9 & 76.5 & 67.7 & 93.5 & 91.1 & 86.1 & 95.3 & 90.4 & 94.3 \\
+WD & \textbf{90.5} & 82.8 & 77.2 & 68.1 & \textbf{94.5} & \textbf{92.0} & 86.9 & \textbf{95.8} & 90.9 & 94.5 \\
+CLICC & 90.1 & 82.7 & 77.3 & 68.2 & 94.1 & 91.7 & 86.8 &95.7 & 90.8 & 94.5 \\
+PCL & 89.8 & \textbf{83.2} & \textbf{78.2} & \textbf{69.0} & 93.7 & 91.5 & \textbf{87.0} & 95.6 & \textbf{91.0} & \textbf{94.6} \\
\bottomrule
\end{tabular}
\caption{Results on Rewrite-20k. All models are initialized from pretrained Bert-base-Chinese model. All results are extracted from the original papers except RUN. For RUN, we reproduce the results from released code to ensure a fair comparison as we are adding modules on RUN. The final line is the result of our complete model equipped with all three modules.}
\label{tb:rewriteresult}
\end{table*}

Our results on MULTI\cite{pan-etal-2019-improving} and REWRITE \cite{su-etal-2019-improving} are shown in Table  \ref{tb:multiresult} and Table \ref{tb:rewriteresult} respectively. On both datasets, our model equipped with one extra module (WD) already surpasses the existing best model on all metrics. The full model equipped with all three modules largely improve the overall performance on each metric. 

Observing from the ablation results of three modules, the F-score performance would drop a bit while BLEU and ROUGE are better after adding CLICC. Through manually analyzing the predicted utterance before and after adding CLICC, we find that CLICC helps reduce the repetition of keywords in generated utterances. This could improve the fluidity and correctness of the sentence while restraining the ability of WD module, which causes the drop of F-score. However, we argue that this feature is useful as we can adjust the importance of two modules for different tasks. For example, the downstream tasks such as FAQ in the dialogue system pay more attention to the recovered keywords would benefit more from WD. In the meantime, if we would show the rewritten queries to users, a more fluent and correct utterance may fit better. Finally, adding PCL boosts the performance again.

\subsection{Influence of Temperature}
\begin{figure}[htb]
\large
\centering
\includegraphics[scale=0.4]{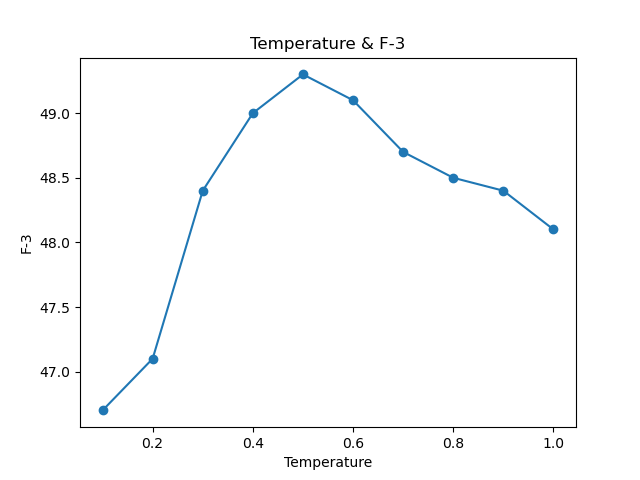}
\caption{The relation between temperature and F3. We get best F3 when temperature is 0.5.}
\label{pic:TEMP}
\end{figure}

The temperature $\tau$ in NT-Xent loss Equation \ref{eq:lossicon} is used to control the smoothness of the distribution normalized by softmax operation. A large temperature smooths the distribution while a small temperature sharpens the distribution. A smoother distribution is easier to learn while risking being not discriminative enough. We explore the influence of temperature in Figure \ref{pic:TEMP}. 
The performance is sensitive to the temperature. A unsuitable temperature will degrade the model  performance. The optimal temperature is obtained around 0.5. This phenomenon demonstrates that, as most negative sentences are far to each other (naturally semantically different), a small temperature may make this task too hard to learn since the model should learn a more general difference rather than detailed differences between anchor and negative samples. A too large temperature is also inappropriate as it may hide the general differences among samples.

\subsection{Different negative strategies}
How to construct hard semantically negative samples is critical for CLICC. One can naturally come up with two ideas: a) Since anchor is semantically complete, we can use original incomplete query as the negative sample.  b) Erasing the longest informative common span between context and current query, which could corrupt the query into a incomplete or more incomplete one. These two strategies are natural as they follow the intention of CLICC which is to force the intent consistency. However, a common disadvantage of these two methods is that they are consistent through the training, i.e. the content of hard semantically negative samples for a certain  data $i$ would not change in the training process. This results in a relatively stagnation in local optimal for the representation of intent. Thus, we adopt the random deletion as our final strategy. The comparison of performance on Rewrite dataset is shown in Table \ref{tb:negstrategies}.

\begin{table}[htb]
\centering
\setlength{\tabcolsep}{2mm}{
\begin{tabular}{c | cccc}
\toprule
Strategies  & $F_3$ & \textbf{EM} &  \textbf{B}$_4$ & \textbf{R}$_L$ \\
\midrule
ORIGIN & 76.7 & 67.0 & 86.8 & 94.3\\
SE & 76.4 & 68.0 & 86.7 & 94.0 \\
RD & 77.3 & 69.0 & 87.0 & 94.6 \\
\bottomrule
\end{tabular}
}

\caption{Reuslts on Rewrite-20k with different negative strategies. ORIGIN means we use original incomplete query as negative samples. SE represents Span Erase, which means we erase the commmon span bettwen context and current query. RD means random deletion which is explained in section about CLICC.}
\label{tb:negstrategies}
\end{table}

\section{Conclusion}
In this work, we explore the key traits of utterance rewriting. We adopt contrastive learning method to model the intent consistency at sentence level and probability consistence in probability space. With the help of carefully designed  combination of multi-tasks, our approach achieves the best performance on several public datasets. In the future, we will  explore modeling more effective positive and negative samples in contrastive learning to improve the utterance rewriting.

\bibliography{main}
\bibliographystyle{acl_natbib}

%
%

\end{document}